\documentclass[sigconf, authorversion]{acmart}

\usepackage{comment}    
\usepackage{graphicx}  
\usepackage{caption}
\usepackage{subcaption}
\usepackage{caption}
\usepackage[utf8]{inputenc}
\usepackage{multicol}
\usepackage{mathrsfs}
\usepackage{mathtools}   
\usepackage{cancel}

\usepackage{hhline}
\usepackage{tabularx,colortbl}
\usepackage{comment}
\usepackage{float}
\usepackage{multirow}
\usepackage{booktabs}
\usepackage{flushend}
\usepackage{algorithm}
\usepackage{bm}
\usepackage{algorithm}
\usepackage{algpseudocode}
\usepackage{algpseudocode}
\usepackage{comment}

\def\cA{\mathcal A}

\def\cC{\mathcal C}

\def\cT{\mathcal T}

\newcommand{\hide}[1]{}
\newcommand{\raf}[1]{(\ref{#1})}

\newcommand{\abs}[1]{\ensuremath{\left|#1\right|}}

\DeclareCaptionType{copyrightbox}

\let\existstemp\exists
\let\foralltemp\forall
\renewcommand*{\exists}{\existstemp\mkern5mu}
\renewcommand*{\forall}{\foralltemp\mkern5mu}

\makeatletter
\newcommand*\bigcdot{\mathpalette\bigcdot@{.5}}
\newcommand*\bigcdot@[2]{\mathbin{\vcenter{\hbox{\scalebox{#2}{$\m@th#1\bullet$}}}}}
\makeatother

\copyrightyear{2023}
\acmYear{2023}
\setcopyright{cc}
\acmConference[e-Energy '23]{The 14th ACM International Conference on Future Energy Systems}{June 20--23, 2023}{Orlando, FL, USA}
\acmBooktitle{The 14th ACM International Conference on Future Energy Systems (e-Energy '23), June 20--23, 2023, Orlando, FL, USA}
\acmDOI{10.1145/3575813.3595205}
\acmISBN{979-8-4007-0032-3/23/06}

\begin{document}
\title{\textsc{DClEVerNet}: Deep Combinatorial Learning for Efficient EV Charging Scheduling in Large-scale Networked Facilities}

\author{Bushra Alshehhi}
\authornote{Both authors contributed equally to this work.}
\email{100050085@ku.ac.ae}
\affiliation{%
  \institution{Khalifa University}
  \city{Abu Dhabi, P.O. Box 127788}
  \country{United Arab Emirates}
}
\author{Areg Karapetyan}
\email{areg.karapetyan@nyu.edu}
\affiliation{%
  \institution{Khalifa University}
  \city{Abu Dhabi, P.O. Box 127788}
  \country{United Arab Emirates}
}
\affiliation{%
  \institution{New York University Abu Dhabi}
  \city{Abu Dhabi, P.O. Box 129188}
  \country{United Arab Emirates}
}
\authornotemark[1]

\author{Khaled Elbassioni}
\email{khaled.elbassioni@ku.ac.ae}
\affiliation{%
  \institution{Khalifa University}
  \city{Abu Dhabi, P.O. Box 127788}
  \country{United Arab Emirates}
}

\author{Sid Chi-Kin Chau}
\email{sid.chau@acm.org}
\affiliation{%
  \institution{Australian National University}
  \city{Canberra ACT 2601}
  \country{Australia}
}

\author{Majid Khonji}
\email{majid.khonji@ku.ac.ae}
\affiliation{%
  \institution{Khalifa University}
  \city{Abu Dhabi, P.O. Box 127788}
  \country{United Arab Emirates}
}

%\affiliation{Division of Engineering, New York University Abu Dhabi, Abu Dhabi, United Arab Emirates}

%\author{Bushra Ali Alshehhi, Areg Karapetyan*, Majid Khonji, Khaled Elbassioni, Sid Chi-Kin Chau} 
\thanks{This work was supported by the Khalifa University of Science and Technology under Award No. CIRA-2020-286.}

\renewcommand{\shortauthors}{Alshehhi and Karapetyan, et al.}

\markboth{}%
{}




\begin{abstract}
With the electrification of transportation, the rising uptake of electric vehicles (EVs)
might stress distribution networks significantly, leaving their performance degraded and stability jeopardized. To accommodate these new loads cost-effectively, modern power grids require coordinated or ``smart'' charging strategies capable of optimizing EV charging scheduling in a scalable and efficient fashion. With this in view, the present work focuses on reservation management programs for large-scale, networked EV charging stations. We formulate \textit{a time-coupled binary optimization} problem that maximizes EV users' total welfare gain while accounting for the network's available power capacity and stations' occupancy limits. To tackle the problem at scale while retaining high solution quality, a data-driven optimization framework combining techniques from the fields of Deep Learning and Approximation Algorithms is introduced. The framework's key ingredient is a novel input-output processing scheme for neural networks that allows \textit{direct extrapolation} to problem sizes substantially larger than those included in the training set. Extensive numerical simulations based on synthetic and real-world data traces verify the effectiveness and superiority of the presented approach over two representative scheduling algorithms. Lastly, we round up the contributions by listing several immediate extensions to the proposed framework and outlining the prospects for further exploration.

\end{abstract}

\keywords{Electric Vehicles, Charging Scheduling, Deep Neural Networks, Algorithm Design, Combinatorial Optimization, Smart Grid.}

\begin{CCSXML}
<ccs2012>
   <concept>
       <concept_id>10003752.10003809.10003636.10003810</concept_id>
       <concept_desc>Theory of computation~Packing and covering problems</concept_desc>
       <concept_significance>500</concept_significance>
       </concept>
   <concept>
       <concept_id>10003752.10003809.10011254</concept_id>
       <concept_desc>Theory of computation~Algorithm design techniques</concept_desc>
       <concept_significance>500</concept_significance>
       </concept>
   <concept>
       <concept_id>10010583.10010662.10010668.10010672</concept_id>
       <concept_desc>Hardware~Smart grid</concept_desc>
       <concept_significance>500</concept_significance>
       </concept>
   <concept>
       <concept_id>10003752.10003809.10003716.10011136</concept_id>
       <concept_desc>Theory of computation~Discrete optimization</concept_desc>
       <concept_significance>500</concept_significance>
       </concept>
   <concept>
       <concept_id>10010147.10010257.10010293.10010294</concept_id>
       <concept_desc>Computing methodologies~Neural networks</concept_desc>
       <concept_significance>500</concept_significance>
       </concept>
   <concept>
       <concept_id>10003752.10003809.10003636.10003808</concept_id>
       <concept_desc>Theory of computation~Scheduling algorithms</concept_desc>
       <concept_significance>500</concept_significance>
       </concept>
 </ccs2012>
\end{CCSXML}

\ccsdesc[500]{Theory of computation~Packing and covering problems}
\ccsdesc[500]{Theory of computation~Algorithm design techniques}
\ccsdesc[500]{Hardware~Smart grid}
\ccsdesc[500]{Theory of computation~Discrete optimization}
\ccsdesc[500]{Computing methodologies~Neural networks}
\ccsdesc[500]{Theory of computation~Scheduling algorithms}

\maketitle

% \fi

%\thispagestyle{plain}
%\pagestyle{plain}
\section{Introduction}

Recent years have witnessed a surge in the adoption of electric vehicles (EVs) as eco-conscious and economical alternatives to combustion engine vehicles. Case in point, over one million EVs were sold in China in $2018$ alone~\cite{Yang2019}, and it is estimated that by $2040$, approximately $700$ million EVs will hit the roads worldwide~\cite{evout}. Smart Grid (SG) technologies can facilitate EV integration and allow utilizing their elasticity for real-time load regulation purposes. However, even with a moderate EV population, the excess demand for charging power might place local distribution circuits under critical strain, leading to potential stability issues and deteriorated efficacy~\cite{5471115, 5356176}. In fact, as demonstrated in~\cite{5471115, 5356176}, just $10\%$ EV penetration rate in a residential distribution network suffices to cause inordinate voltage deviations, branch congestion, and overheating of substation transformers, leaving the power system equipment with shortened lifespan.

While the increased energy demand can be supported via gradual capacity upgrades, the sudden spikes in EV and non-EV loads require expensive fast generators as a backup, entailing high power losses. Coordinated, or ``smart'', charging strategies can mitigate these issues by exploiting the flexibility of EV users' charging reservation requests (e.g., at which station and/or time period to charge). By optimizing the charging schedule based on these requests, coordinated charging can improve energy utilization while still meeting the needs of EV owners. Such schemes can be deployed in public parking sites, workplaces, shopping malls, residential complexes and would be particularly effective for large-scale charging facilities.

To unlock this tremendous demand-side management potential, SG operators require \textit{efficient} and \textit{scalable} means to tackle large-scale scheduling problems. The optimization involved is time-coupled and \textit{combinatorial} in nature: deciding which EV may charge at which station/rate and time given the available net power supply over the scheduling horizon. Unlike convex programming, which can be readily handled by off-the-shelf numerical solvers (e.g., Cplex or Gurobi), combinatorial (discrete) optimization is generally notorious for being computationally expensive. To exemplify, Fig.~\ref{fig:rnt} plots the execution time of Gurobi when applied to the combinatorial EV charging reservation management problem at hand (dubbed {\sc EVCRP} and formalized in Sec.~\ref{sec:problem}). As observed from Fig.~\ref{fig:rnt}, with growing problem size, the solver's running time quickly turns prohibitive and may vary significantly depending on the input data. 

Thanks to the inherent massive parallelism and powerful learning capacity (of approximating sophisticated non-linear mappings from labeled data), Deep Neural Networks (DNNs) have gained traction as an increasingly viable method to streamline decision-making in complex large-scale engineering systems~\cite{goodfellow2016deep, LOPEZGARCIA2020103894, HORNIK1991251}. Specifically, problem instances routinely solved in practice often share similar patterns or stem from related data distributions, which DNNs can exploit and learn to imitate the known optimal/near-optimal iterative algorithms, thereby dramatically boosting the computing speed. However, the application of DNNs to constrained optimization problems confronts with a number of hurdles. First, the solutions predicted by DNNs may violate the problem constraints. Second, since the number of neurons in DNNs' input layer is normally predetermined, even a tiny increase in the problem size (e.g., when several new EV users subscribe to {\sc EVCRP} program) will necessitate retraining the model\footnote{When the input data is straightforwardly fed into DNN.}. Last but not least, obtaining training sets for large-scale combinatorial problems with tens of thousands of decision variables could prove computationally intractable unless one accepts crude approximations.

To address the aforementioned concerns, we introduce a Deep Combinatorial Learning framework, coined as \textsc{DClEVerNet}, that can produce \textit{close-to-optimal feasible} solutions to large-scale {\sc EVCRP} instances in sub/near-linear time. Notably, the proposed approach conforms to {\sc EVCRP} \textit{inputs of arbitrary cardinality} without requiring to retrain or alter the network structure (namely, the number of neurons and connecting weights) for each input size, while also maintaining satisfactory performance as the count of participant EVs continues to rise. In summary, the current work complements and advances the existing research with the following three-fold contributions:

 \begin{figure}[!t]
     \centering
     \includegraphics[trim={0.2cm 0cm 0.2cm 0cm},clip, width=\columnwidth]{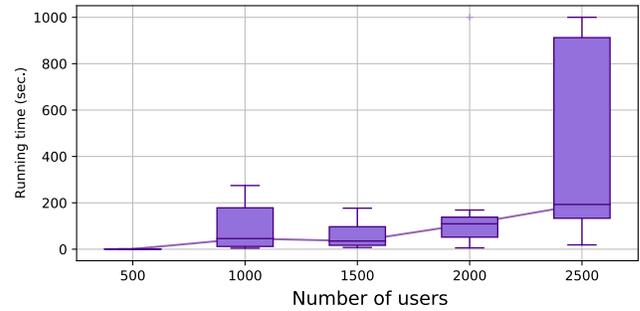}
     \caption{Average computational time (across $10$ runs) taken to solve randomly generated {\sc EVCRP} instances (considering $3$ charging stations) with the Gurobi optimizer against the input size at 95\% confidence interval. Outliers are plotted as crosses.}
     \label{fig:rnt}
 \end{figure}
\begin{enumerate}
    \item By drawing on ideas from the toolbox of algorithmic techniques, we devise a computationally conducive input-output (IO) representation scheme for neural networks allowing to tackle packing (and covering) combinatorial problems in a \textit{scalable} and \textit{efficient} fashion. With this scheme in place, a neural network trained on small problem sizes can be \textit{directly invoked} (without retraining) to generate approximate solutions for the problem in high dimensional spaces. Furthermore, the adaptive structure of the scheme allows tuning the \textit{granularity} of predicted optimization decisions to govern the trade-off between complexity and performance of the learning process (in a sense resembling an approximation scheme). 
    \item We assemble a sample \textsc{DClEVerNet} model\footnote{The trained model along with its sample application tutorial can be accessed online at \url{https://drive.google.com/drive/folders/17miO6eaIxDqYmtXGrPQHV1brnjU81owh}.} consisting of a feed-forward DNN\footnote{Note that the framework is applicable beyond DNNs and can incorporate more complex architectures, such as Convolutional Neural Networks or Transformers as elaborated in Sec.~\ref{sec:discussion}.} augmented by the proposed IO scheme and complemented by a simple (sorting based) and fast post-processing routine for extracting feasible, approximate solutions to {\sc EVCRP}. Leveraging the scalability of \textsc{DClEVerNet}, we employ the model trained on the inputs of {\sc EVCRP} problem for $1500$ users ($4500$ binary decision variables) to produce solutions to instances with up to $10,000$ users ($30,000$ binary decision variables), achieving on
    average \textit{nearly $80$\%} of the optimal objective value. 

    \item To consolidate the contributions, the performance of the featured approach is scrutinized extensively under various case studies based on both synthetic and \textit{real-world data traces} (acquired from Caltech's Adaptive Charging Network (ACN)~\cite{acn_dataset}). First, we demonstrate the scalability and efficiency of \textsc{DClEVerNet} by contrasting it against two benchmarks: (i) An adapted version of the polynomial-time approximation scheme (PTAS) recently developed in~\cite{khonji2018approximation}; (ii) A greedy baseline strategy prioritizing higher-valued requests. Subsequently, we investigate the generalisability and robustness of \textsc{DClEVerNet} to out-of-sample and out-of-distribution inputs generated by perturbing the values of {\sc EVCRP} parameters (e.g., the available power capacity or the number of charging slots in stations).

\end{enumerate}

The remainder of this paper is organized as follows. Sec.~\ref{sec:related} reviews the related literature on the EV charging scheduling problem. Sec.~\ref{sec:problem} formulates the problem mathematically. In Sec.~\ref{sec:dnn}, we lay out the proposed Deep Combinatorial Learning approach, and in  Sec.~\ref{sec:sims} validate its performance through extensive simulation studies. Lastly,  Sec.~\ref{sec:discussion} sketches several immediate extensions for future work, followed by concluding remarks in Sec.~\ref{sec:conclusion}.

\section{Related Work}\label{sec:related}

As surveyed in~\cite{tan2016integration,amjad2018review,mukherjee2014review}, a considerable body of literature has been published on controlled EV charging, proffering a rich arsenal of techniques to cater for various operational objectives, such as relieving power system congestion~\cite{Zishan10}, maximizing EV owners' convenience~\cite{7460201}, minimizing charging expenses~\cite{8703849}, enhancing voltage profile~\cite{5986769}, valley filling~\cite{7460201, 6313962}, to name a few. From a methodological standpoint, the existing approaches can be broadly categorized into three groups: approximation algorithms, heuristics/meta-heuristics, and Deep Learning (DL) based methods.

Approximation algorithms are relatively simpler and faster than exact solution methods, such as branch-and-bound and dynamic programming. The salience of these algorithms manifests in provable optimality guarantees quantified by \textit{approximation ratio}, which measures the performance gap against the optimal objective value over all possible input realizations (i.e., in the worst possible scenario). In ~\cite{alinia2020online}, Alinia et al. formulate {\sc EVCRP} as a $0$-$1$ programming problem that aims to maximize the total revenue obtained from charged EVs while respecting local and global peak power constraints. Assuming all EVs start charging simultaneously, the authors develop a primal-dual scheduling algorithm with a bounded approximation ratio. Towards a more realistic heterogeneous scenario, Majid et al.~\cite{khonji2018approximation} devised a $(1-\epsilon)$ polynomial-time approximation scheme (PTAS), where $\epsilon \in (0,1)$ parameterizes the desired approximation ratio and the running time, to solve {\sc EVCRP} for constant-length scheduling horizons. Though  theoretically significant, these methods could be of limited practicality as improvements in solution quality typically come at the expense of scalability.

As such, heuristics/meta-heuristics are \textit{devoid of
any} optimality guarantees, yet often prove practically valuable when applied to combinatorial problems. The study in~\cite{8039201} formulates {\sc EVCRP} as a multi-objective optimization problem to maximize EV users' revenue (by minimizing the charging cost and duration while also increasing the supplied energy). An evolutionary-inspired heuristic search algorithm, Ant Colony, is invoked to attain scalability. However, the setting in~\cite{8039201} does not account for the power network capacity constraint. Taking a step further, Liu et al.~\cite{liu2019coordinated} consider an extended formulation incorporating the power constraints. Yet, a somewhat restrictive assumption is imposed that EVs must stop in multiple charging stations to fulfill their power demand. On the other hand, Sun et al. ~\cite{7460201} frame {\sc EVCRP} as two consecutive problems, where the first schedules the requests to achieve a flat load profile, whereas the second, which is combinatorial, aims to minimize interruptions during charging to preserve EVs' battery life. The authors provide a simple heuristic algorithm and evaluate its performance empirically through simulations.

Different from the above two, DL-based approaches can learn representations of data and then capitalize on the inferred information to guide or aid the decision-making process. In general, the existing learning-based approaches to optimization problems fall under two themes: end-to-end (i.e., standalone) and hybrid. Studies focusing on the former, e.g., ~\cite{pan2020deepopf,dobbe2019toward,sanseverino2016multi,zhao2020deepopf+}, utilize DL techniques 
to predict the solutions directly (without solving the optimization problem) by mapping the input data to the desired output (e.g., the optimal solution). For instance, the work in~\cite{zhao2020deepopf+} constructs a DNN to learn the optimal load-generation mapping in the DC Optimal Power Flow problem. Such methods can capture complex relationships between the input data and the output, yet may not be able to exploit the problem's structural properties. In contrast, hybrid approaches pair known traditional optimization methods or numerical solvers with machine learning techniques~\cite{dong2020smart,gutierrez2010neural,vaccaro2016knowledge, Xu2020, NEURIPS2019_d14c2267}. This allows incorporating problem-specific knowledge/techniques to improve the computational efficiency of these algorithms or speed up the solvers. For example, Xu et al. \cite{Xu2020} empirically enhance dynamic programming with DNNs to tackle combinatorial optimization problems in a computationally more efficient manner. Gasse et al. \cite{NEURIPS2019_d14c2267}, on the other hand, present a Graph Convolutional Neural Network model to speed up mixed-integer linear programming solvers by automatically learning on which variables to branch.

Among various DL techniques, DNNs~\cite{LeCun2015} are by far the most prevalent and mature models~\cite{goodfellow2016deep}. Typically, DNNs comprise a series of layers stacked on
top of each other, which collectively seek to approximate complex non-linear mappings. The universal approximation capability~\cite{HORNIK1991251} coupled with the coveted computational efficiency renders DNNs particularly suited for large-scale optimization tasks. This stimulated application of DNNs to smart charging of EVs~\cite{8299470, 8910361,aljafari2023electric,shibl2020machine,shibl2021electric} (see~\cite{9469715} for a recent review) as well as to power system problems in general~\cite{zhao2020deepopf+, Fioretto_Mak_Van}. The study in~\cite{aljafari2023electric} proposed to utilize Deep Reinforcement Learning within a dynamic pricing framework as an alternative to the traditional on-peak and off-peak pricing. Some prior research studies considered a simplified variant of {\sc EVCRP}, where the problem is reduced to a classification task of predicting the EV charging rate, charging location, or the mode of charging (i.e., charging or discharging)~\cite{shibl2020machine,shibl2021electric}. In ~\cite{shibl2021electric}, the authors compare the performance of various machine learning approaches, including Decision Tree (DT), Random Forest (RF), Support Vector Machine (SVM), K-Nearest Neighbours (KNN), Long Short-Term Memory (LSTM) and DNN. The developed method relies on two separate classifiers for predicting the charging rate and the charging location, respectively. For both classifiers, RF and LSTM achieved the highest accuracy. However, it is important to note that casting {\sc EVCRP} as the said classification task limits the network operator functionalities, hence the demand-side management potential.

To reduce charging costs, Lopez et al.~\cite{8299470} develop a DNN-based strategy that can determine the optimal EV charging periods in response to real-time electricity price signals. The model was trained with historical data of charging sessions as inputs and the corresponding optimal solutions (calculated by Dynamic Programming) as prediction labels. In~\cite{8910361}, Li et al. additionally consider the effect of discharging operations and propose a reinforcement learning framework wherein policies are optimized by invoking DNNs. Nevertheless, for constrained optimization problems, which are a departure from those studied in~\cite{8299470, 8910361}, DNNs \textit{do not necessarily} guarantee a feasible solution due to lacking representation of the exact feasibility region as well as their inherent and inevitable approximation errors. Recently, Zhao et al.~\cite{zhao2020deepopf+} proposed a DNN-based scheme for DC Optimal Power Flow problem that assures the feasibility of output generator set-points. The underlying idea is to ``preventively'' adjust (perturb by an appropriate magnitude) the constraints during the training stage, thereby anticipating approximation errors and warding off potential violations in the testing phase. While beneficial on its own, this method does not seem amenable to combinatorial optimization problems with binary/integer variables.

\section{Problem Statement}\label{sec:problem}

Recall that in the problem under study, the charging network operator seeks to determine binary scheduling assignments (reflecting the accept/reject decisions) for the charging reservation requests of subscribed EV users containing the preferred charging stations (CSs), intervals, and valuations. We assume a centralized control scheme wherein the requests are elicited in an apriori manner (e.g., a day ahead); hence, all users' profiles are accessible beforehand. 

In the said network, embedded within and supplied by a power distribution grid, CSs are indexed by the set $\cC$ and distinguished by their charging power rate $r_c \in \mathbb{R}, c \in \cC$. We consider a time-slotted system model in which the scheduling horizon $\cT \triangleq \{1,\hdots, T\}$ is discretized into $T$ equidistant intervals according to the desired frequency of control signals. In the set of participant EVs, denoted by $\cA$, each customer $a \in \cA$ lodges a charging request which includes $\Big(\cC^a, \{\mathcal{T}^a_c\}_{c \in \cC^a}\Big)$, where $\cC^a \subseteq \cC$ is the set of favored CSs and $\mathcal{T}^a_c \subseteq \mathcal{T}$ are the corresponding preferred charging periods, in order to fulfill their charging power demand $P^a = P^a_c \triangleq r_c|\mathcal{T}^a_c|$ for $\forall c \in \cC$ (Note that the power demand is invariant across CSs). Accordingly, shall user $a$'s request be accepted, the scheduler will \textit{select one among} $a$'s preferred options and reserve the corresponding electric vehicle supply equipment (EVSE) for the requested duration. Let $x^a_c$ denote the scheduling decision for user $a \in \cA$, then
$$x^a_c = \begin{cases} 
	1, & \mbox{If user } a\mbox{ is assigned to charge at the CS } c \in \cC^a \\ 
	0 & \text{Otherwise}
\end{cases}\,.$$
To model users' welfare, define the \textit{gain} for each user $a \in \cA$ with respect to the charging decision $x^a_c$ as 
\begin{equation}\label{eqn:gain}
         G^a(x^a_c) \triangleq u^a x^a_c - \sum_{t \in \mathcal{T}^a_c} {\tt{cost}}(t) \cdot x^a_c \,, 
 \end{equation}
where ${\tt{cost}}(t)$ refers to the time-varying electricity cost and $u^a$ is \textit{a user-defined} parameter included in the charging request that measures the \textit{utility} perceived by customer $a$. Here, utility quantifies the extent of user-obtained comfort or, alternatively, the \textit{worthiness (valuation)} of receiving the requested charging power. In this context, $G^a(x^a_c)$ can be interpreted as the savings gained by user $a$ from participating in the reservation program. Without loss of generality, we suppose that $u^a$ is sufficiently large so that $G^a(x^a_c)$ is non-negative for $\forall a \in \cA, c \in \cC$. For brevity, in what follows, we shall write $G^a$ to denote users' gain if the charging request is satisfied (i.e., $x^a_c =1$ for some $c \in \cC^a$) and henceforth refer to $G^a$ as user $a$'s \textit{conditional gain}. Also, we define $R \triangleq \dfrac{G^a}{P^a}$ to be user $a's$ \textit{conditional gain-to-power ratio}.

To account for practical aspects, we assume that in addition to the EV load, the network supplies energy to residential and commercial consumers, referred to as background active power demand or \textit{base load} and denoted by $d(t) \in \mathbb{R}$ at time $t \in \cT$. Additionally, to cater for physical limitations, we bound the total power capacity of the network by $\overline{P}$ and cap the occupancy limit of each CS $c \in \cC$ by $N_c$ corresponding to the number of installed EVSE.

With the above notation, {\sc EVCRP} translates into the following combinatorial optimization problem:
\begin{align}
\displaystyle \max_{x^a_c}\quad &  \sum_{a\in \mathcal{A}}\sum_{c \in \mathcal{C}^a}  G^a (x^a_c) \qquad \qquad \qquad \qquad \qquad \qquad  \textsc{\big(EVCRP\big)} \nonumber\\ 
\text{s.t.}\quad&  d(t) + \sum_{a\in \mathcal{A}}  \sum_{c \in C^a ~:~ t \in \cT^a_c } r_c \cdot x^a_c   \leq \overline{P}, \quad \forall t \in \mathcal{T} \label{eqn:cons-one} \\
&  \sum_{c \in C^a} x^a_c \leq 1, \quad \forall a\in \mathcal{A}\label{eqn:cons-two} \\
& \sum_{a\in \mathcal{A}~:~c\in C^a,~ t\in \cT^a_c} x^a_c \leq N_c ,  \quad \forall c \in C, t \in \cT \label{eqn:cons-three}\\
&x^a_c\in \lbrace{0,1}\rbrace, \quad \forall a \in \mathcal{A}, c \in \cC\,.\label{eqn:cons-four}
\end{align}
{\sc EVCRP} seeks to maximize users' total welfare gain while considering both local and global peak constraints in the networked CSs. Constr.~\raf{eqn:cons-one} stipulates the total supplied power to remain below the network's maximum capacity. Constr.~\raf{eqn:cons-two} ensures that each user is assigned only to one CS. Constr.~\raf{eqn:cons-three} imposes a limit on the number of customers who can charge at a particular CS at any given time so that it does not exceed the maximum number of EVSE. Finally, Constr.~\raf{eqn:cons-four} enforces the integrality of decision variables.

The crux of solving {\sc EVCRP} lies in its combinatorial structure imposed by the binary decision variables $X \triangleq \big(x^a_c\big)_{a \in \cA, c \in \cC}$. As such, {\sc EVCRP} specializes to several classical NP-hard combinatorial problems, including the two-dimensional Knapsack problem (when $|\cT|=|\cC|=1$) and the Unsplittable Flow on a Path problem (when $|\cC|=1$ and Constr.~\raf{eqn:cons-three} is ignored). This rules out {\sc EVCRP}'s NP-hardness and hints that it is substantially more complicated than the latter two problems. Given these facts, designing scalable and near-optimal scheduling algorithms for {\sc EVCRP} becomes markedly challenging. In the proceeding section, through a judicious combination of algorithmic and DL techniques, we devise a Deep Combinatorial Learning framework that, on average, can closely approximate the optimal solutions of large-scale {\sc EVCRP} instances in linear time. 

\section{Proposed Approach}\label{sec:dnn}
This section introduces \textsc{DClEVerNet} and provides a detailed description of its architecture. First, we outline the methodology of the proposed framework and highlight the challenges inherent to designing a learning-based approach for combinatorial optimization problems, and how the framework addresses these hurdles. The framework's structure is comprised of three modules: (1) Pre-processing step that downsamples (compresses) {\sc EVCRP} input into a succinct representation; (2) DNN for predicting the distribution of users in the optimal solution; (3) Fast post-processing procedure to extract a feasible solution from the prediction. Each of these components is thoroughly analyzed and explained in the following subsections.

\paragraph{\textbf{Notational Convention:}} In the remainder of this paper, unless otherwise explicitly stated, we shall enclose the inputs of procedures, such as counting or averaging, within the operator $[~ ]$. Given a vector/set $v$ we let $|v|$ symbolise its magnitude/size and we write $\tilde{v}$ to denote the normalization of the elements in $v$ by a certain value.

\subsection{Overview and Challenges}
\begin{figure}[!b]
     \centering
     \includegraphics[trim={2.2cm 0cm 0cm 0cm},clip,width=\columnwidth]{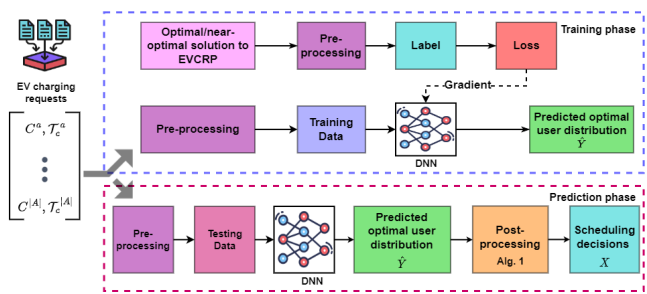}
     \caption{High-level block diagram of the featured DL-assisted optimizer enabling rapid computation of near-optimal solutions to {\sc EVCRP}.}
     \label{fig:PTAS-SETS}
 \end{figure}

As illustrated in~Fig.\ref{fig:PTAS-SETS}, which depicts the high-level schematics of \textsc{DClEVerNet}, the methodology can be broken down into two distinct phases: training and prediction. During the training phase, input instances of {\sc EVCRP} and their corresponding solutions (from which the prediction labels are constructed) are fed to the neural network. The primary objective of the training phase is for the DNN to acquire the ability to predict solutions for previously unseen instances of {\sc EVCRP} during the prediction phase. 

 \begin{figure*}[!t]
     \centering
     \includegraphics[width=\textwidth]{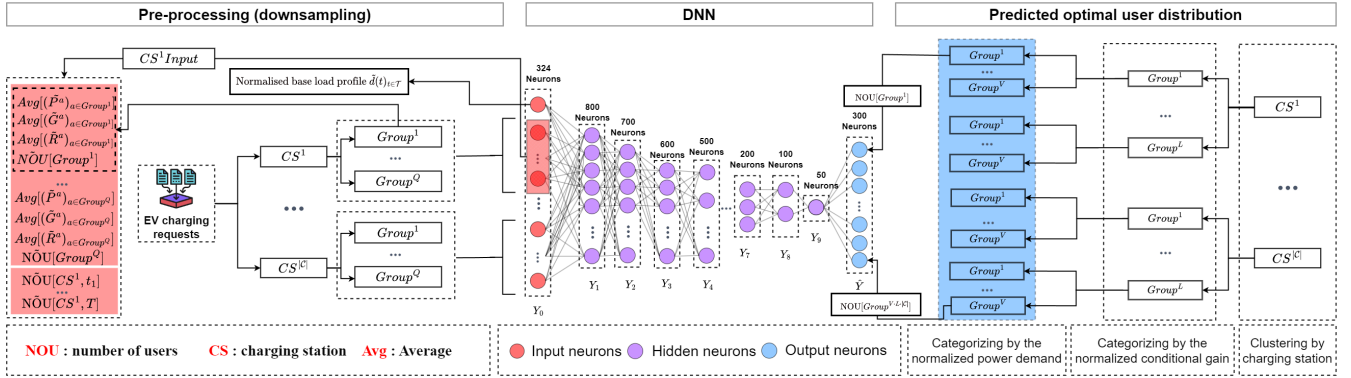}
     \caption{Detailed architecture of the 
     constructed sample \textsc{DClEVerNet} model comprised of a DNN augmented by the proposed input-output processing scheme. For scalability, in the pre-processing step, {\sc EVCRP} input is downsampled via grouping and averaging to allow succinct representation. The output layer predicts the number of users within groups in {\sc EVCRP}'s optimal solution.}
     \label{fig:Deep-Learning-overview}
 \end{figure*}

A significant amount of research has been conducted to explore the application of DNNs for solving combinatorial optimization problems (COPs). One of the challenges encountered is adequately representing the optimization problem as features and labels (i.e., input and output of the DNN). COPs often involve a large number of variables and complex constraints and objectives, making it difficult to encode an appropriate input representation for the network. Despite the robust nature of DNNs, poorly represented input can affect their performance.

A well-defined input reduces the number of parameters to be learned by the network, thus resulting in a more efficient training process. Furthermore, compact inputs can also help reduce overfitting, which is a common problem in DNNs. Overfitting occurs when a model performs well on the training data but poorly on unseen data. This is because the model has learned to fit the noise and redundancy in the training data rather than generalizing to new examples. 

The main challenge in remodeling a COP solution as a label is the discrete format of the data representation. Applying this to deep learning models that work best with continuous data can be challenging. In addition, encoding the optimal solution as the output of a DNN is not a recommended practice due to its inherent sparsity. The optimal solution to many real-world problems is often sparse. In such cases, encoding the solution directly into the output layer of a DNN would result in a large number of parameters, many of which would have little impact on the solution. This would render the model difficult to train and could lead to overfitting.

In view of the above, the authors in ~\cite{8299470} designed a DNN that outputs decisions for a single EV at a time. Given only the current state of the battery, the output is a single variable to indicate whether to admit the user or not. The downside of this approach is that the decisions are (locally) made without considering the entire set of EV requests, which could lead to highly suboptimal global solutions. The study in ~\cite{huang2021deepopf} addresses the issue by utilizing the capability of DNN to predict a partial solution to the optimal power flow problem (OPF). Then, the final solution is constructed by solving OPF initialized with the DNN output. This approach is suitable for small-scale instances since it requires numerical solvers for constructing the final solution and is hence limited by the capabilities of these solvers.

Feeding DNNs with a feasible solution is not sufficient to guarantee the feasibility of output solutions. Therefore, the authors in ~\cite{pan2022deepopf} propose incorporating a penalty term in addition to the mean square error (MSE) loss function. However, it is still possible that the obtained solution will not satisfy all the constraints. Therefore, a post-processing step, in which the nearest feasible solution is chosen greedily, was employed. To ensure feasibility, in ~\cite{zhao2020deepopf+}, the authors improve the approach by calibrating the penalty to be enforced when the constraints are  within a percentage of the violation value. While this approach shows some promise, 
it is not trivial to tune the calibration for optimal performance. In addition, for {\sc EVCRP}, calibration of constraints may lead the network to learn a significantly suboptimal solution due to the discreteness of the problem.

Importantly, all of the aforementioned studies \textit{require re-training} if the input size (e.g., number of EV requests) is varied. The present work introduces a novel approach that overcomes the above limitations. In \textsc{DClEVerNet}, we tackle the encoding of {\sc EVCRP} inputs and outputs through a judiciously constructed scheme (detailed in Sec.~\ref{sec:preprecoess}) that enables \textit{scalability}. Unlike existing methods in the related literature, we model the DNN input as groups of EV charging requests. This parses the input size invariant to the number of EV charging requests. The output of the network is the optimal distribution of allocated users with respect to groups, thus allowing the network to be able to handle large-scale instances. The feasibility is assured through a lightweight post-processing routine that extracts the solution from the predicted distribution (detailed in Sec.~\ref{sec:postprocoess}).

In the current study, we utilize a training dataset labeled with optimal solutions obtained from the Gurobi solver. However, for certain combinatorial optimization problems, such as the Traveling Salesman Problem, it may be computationally infeasible to find the optimal solution when the problem size is large. An alternative approach is to construct the training data based on near-optimal solutions obtained from known efficient approximation algorithms. By training the model on these solutions, the network can learn to imitate the underlying principles of the utilized approximation algorithm, thereby unlocking substantial speedup gains.

\subsection{Pre-processing}\label{sec:preprecoess}

Recall that {\sc EVCRP}'s input consists of EV users' charging reservation requests characterized by the set of favored CSs, the corresponding preferred charging intervals, and conditional gains associated with each request. As seen from Fig.~\ref{fig:Deep-Learning-overview}, we first partition the requests by the requested charging station into $|\cC|$ groups. For each station $c \in \cC$, the conditional gains for users in the respective group are normalized by dividing by the maximum value within that group. Then, the requests in each group are further classified into $Q$ groups based on their normalized conditional gains in descending order. In particular, the set of users $q_i$ in the $i^{th}$ group is as follows:
\begin{equation}\label{eqn:groups}
    q_i \triangleq \Bigg\lbrace{a :  \frac{i-1}{Q}< \tilde{G}^a \leq \frac{i}{Q}, a \in \mathcal{A}} \Bigg\rbrace, \quad \forall ~~ i=1, \hdots, Q\,,
\end{equation}
where $\tilde{G}^a \triangleq \dfrac{G^a}{\max_{a \in q_i} G^a}$ denotes the conditional gain of user $a$ normalized by the maximum conditional gain among the users in $q_i$.
The rationale underlying this clustering criterion is to produce groups wherein the difference between users' \textit{conditional gains is bounded}. In other terms, the users within a group share roughly ``similar'' conditional gains and can be represented by a ``centroid'' user. This, in turn, allows for decision-making at a group level rather than at an individual level without inflicting a significant loss in the total gain of the selected customers. Such grouping schemes have been frequently employed in theoretical computer science literature to devise approximation algorithms for packing optimization problems. For instance, a similar method in which the users are discerned into a logarithmic number of groups wherein their utilities differ by no more than a constant factor was employed in~\cite{Karapetyan2023, kkpl, Elbassioni2019} to devise approximation algorithms for different variants of the Unsplittable Flow on Paths and $r$-weighted Minimization Knapsack problems. This method provides a systematic and efficient manner of downscaling the inputs of packing/covering COPs. Note that the partitioning criteria can be adapted based on the problem type. In addition to conditional gains, users can be alternatively clustered based on their utility, power demand, or utility-to-demand ratio.

For DNN to learn the optimal scheduling of EV charging requests effectively, the network must gain a comprehensive understanding of the 
constraints and parameters of {\sc EVCRP}. In this regard, as illustrated in Fig.~\ref{fig:Deep-Learning-overview} we include the normalized load profile $\tilde{d}(t)_{t \in \cT}$ in the input layer. This intends to convey the congestion status of the power network at any given time slot. Each group $q_i$ is represented by four key input entities that summarize the information of users in $q_i$: 
\begin{enumerate}
    \item $Avg [(\tilde{P}^a)_{a \in q_i}]$ - The average $\tilde{P}^a \triangleq \dfrac{P^a}{\overline{P}}$ of $~\forall a \in q_i$, where $\tilde{P}^a$ is user $a$'s power demand normalized by the capacity of the power network $\overline{P}$.
    \item $\tilde{NOU} [q_i]$ - The group size normalised by the total number of users $|\mathcal{A}|$, i.e., $\tilde{NOU} [q_i] \triangleq \dfrac{|q_i|}{|\cA|}$; 
    \item $Avg [(\tilde{G}^a)_{a \in q_i}]$ - The average $\tilde{G}^a$ (as defined above) of all the users in group $q_i$;
    \item $Avg [(\tilde{R}^a)_{a \in q_i}]$ - The average $\tilde{R}^a \triangleq \dfrac{R^a}{\max_{a \in (q_i)_{i=1,...,Q}}R^a}$ of the users in $q_i$, where  $\tilde{R}^a$ is user $a$'s conditional gain-to-power ratio normalized by the maximum ratio among all the users in the corresponding CS.
\end{enumerate}

In addition to the aforementioned input entities, we also include for each charging station $c \in \cC$ the normalized number of charging requests in each time slot. The normalization was done with respect to the maximum number of requests at any time slot in the corresponding CS. The adopted DNN can benefit from this information by learning the temporal dynamics of the charging requests, which can aid in identifying the peak charging periods. We note that input normalization can play a crucial role in the efficiency of the learning process as well as aid in preventing overfitting. As transpires from the simulation results presented in Sec.~\ref{sec:sims}, taken together, these input entities allowed the trained sample \textsc{DClEVerNet} model to extract transferable knowledge about the structure and parameters of {\sc EVCRP}.

One of the key design considerations when encoding the EV requests as proposed is determining the number of groups $Q$. The latter corresponds to the input data's granularity; thus, increasing the number of groups will result in a more detailed representation of the input. 
This can be beneficial for applications for which accuracy is of crucial importance. However, to ensure efficient training, the size of the training data should be increased. Conversely, decreasing the number of groups compresses the input resulting in a more general representation of the input. The advantage of this is that the DNN will identify the most salient features of the input data, and the compressed representation will highlight the most important information. Aside from the representation, the number of groups governs the number of parameters in the model to be trained, hence directly affecting the required computational resources. Therefore, the input dimensionality should be adjusted considering the trade-off between the desired accuracy and the available computational resources. 

\subsection{Network Architecture}
As illustrated in ~Fig.\ref{fig:Deep-Learning-overview}, we adopt a multi-layer feed-forward neural network architecture of the form:
\begin{eqnarray}
 Y_0& =& \Big[\tilde{d}(t)_{t \in \cT}, ~
 \chi^{CS^1}, \hdots,~\chi^{CS^{|\cC|}}\Big], \label{eqn:FFN1}\\
     Y_i &=& \sigma\Big(W_{i}Y_{i-1}+b_i\Big), \quad \forall~ i=1, 2, \hdots , H \label{eqn:FFN2}\\
    \hat{Y}&=&\sigma^{\prime}\Big(W_{H+1}Y_{H}+b_{H+1}\Big), \label{eqn:FFN3}
\end{eqnarray}
where $\chi^{CS^1}$ encodes the input entities for CS $1$, as depicted in Fig.~\ref{fig:Deep-Learning-overview} and detailed above, $Y_0$ and $\hat{Y}$ are the input and output layers, respectively, while $Y_i$ is the output vector of the $i$-th hidden layer (out of $H$ in total) which depends on the weights $W_i$, biases $b_i$, and the output of the previous layer $Y_{i-1}$. In Eqns.~\raf{eqn:FFN2} and~\raf{eqn:FFN3}, $\sigma(.)$  and $\sigma^{\prime}(.)$ stand for the Rectified Linear Unit (ReLU) activation functions used in the hidden layers and the linear activation function for the output layer, respectively. ReLU  allows the network to learn non-linear relationships in the data, and the linear activation function allows the output layer to produce  continuous values that represent the number of users in each group. In the present study, we construct a DNN with nine hidden layers, thus $H=9$. The number of neurons in the first hidden layer is $800$, gradually decreasing to $50$ as the layers become deeper. As a loss function, we utilize the mean squared error (MSE):
\begin{equation}
  \mathcal{L}_{MSE} = \frac{1}{n} \sum_{i=1}^{n} \big(Y^i - \hat{Y}^i\big)^2 \, ,
  \end{equation}
where $n$ is the number of samples, $\hat{Y}^i$ is the prediction of the network for sample $i$, and $Y^i$ is the corresponding ground truth value. The selection of MSE as a loss function rests on extensive experimentation with various alternatives, including Huber Loss, mean percentage error, and mean absolute error. In particular, the experimentation results revealed that MSE provided the most accurate and reliable performance.

Similar to the grouping in the input layer, the EV requests in the output are grouped  by the requested charging station into $|\cC|$ groups. For each station $c \in \cC$, the requests are partitioned into $L$ groups based on $\tilde{G}^a$ in descending order. Let $l_i$ be the set of users in the $i$-th group, then
\begin{equation}\label{eqn:groups}
    l_i \triangleq \Big\lbrace{a :  \frac{i-1}{L}< \tilde{G}^a \leq \frac{i}{L}, a \in \mathcal{A}} \Big\rbrace, \quad \forall ~~ i=1, \hdots, L\,.
\end{equation}
For further granularity, in each group $l_i$, the power demand is normalized by dividing over ~$\overline{P}$, and the requests are  partitioned into $V$ groups based on their normalized power demand in ascending order. Denote by $v_i$ the set of users in the $i$-th group such that
\begin{equation}\label{eqn:groups}
    q_i \triangleq \Big\lbrace{a :  (1- \frac{i}{V})< \tilde{P}^a \leq \frac{V-i+1}{V}, a \in \mathcal{A}} \Big\rbrace, \quad \forall ~~ i=1, \hdots, V\,,
\end{equation}
where $\tilde{P}^a$  denotes the normalized power demand of user $a \in \mathcal{A}$. 

The groups have been arranged so that the first group comprises users with the highest conditional gain and the lowest power demand. The output layer denoted by $\hat{Y}$ consists of $\abs{\hat{Y}}=\big(|\cC|\cdot L \cdot V\big)$ neurons. Each output neuron in $\hat{Y}$  represents the number of users to include in the final solution from each group.

\subsection{Post-processing}\label{sec:postprocoess}
As previously noted, the network learns the distribution of user categories in the optimal solution thus requiring post-processing to extract the final solution of {\sc EVCRP}. For the purposes of the current study we employ a greedy approach explained in~Alg. \ref{alg:cap}.

The output of the DNN, $\hat{Y}$, representing the number of users to be assigned to each group $j$, serves as an input to Alg. \ref{alg:cap}. The algorithm proceeds by rounding down the elements of $\hat{Y}$ to their nearest integer values. For each group $j$, we define $\mathcal{M}$ as the set of prospective users in group $j$ sorted in descending order of their $G^a$. Then, the algorithm iterates over $\mathcal{M}$, selecting the users unless the solution is deemed infeasible. The feasibility of the solution is contingent upon constraints ~(\ref{eqn:cons-one}) to ~(\ref{eqn:cons-four}). Constr.~(\ref{eqn:cons-one}) is violated if the maximum network capacity is exceeded at any point in time. Constr. (\ref{eqn:cons-two}) requires a user to be assigned to only one charging station. Constr.(\ref{eqn:cons-three}) is violated if the number of assigned users in any station exceeds $N_c$. Lastly, as the procedure either accepts or rejects users, constraint (\ref{eqn:cons-four}) is ensured to be met. In the event that incorporating a user leads to an infeasible solution, their charging request will be declined, and the algorithm continues to the next user in $\mathcal{M}$. The algorithm terminates when either the proposed number of users to be added to group $j$  is zero or all the users in the group have been evaluated.

\begin{algorithm}[!h]
\caption{Post-processing Routine}\label{alg:cap}
 \hspace*{-200pt}\textbf{Input: $\hat{Y}$} \\
  \hspace*{-122pt}\textbf{Output:} Solution $X$ to {\sc EVCRP}
\begin{algorithmic}[1]
\State { $\hat{Y}=\left\lfloor \hat{Y} \right\rfloor$}
\For {$j\in \hat{Y}$}
    \State { $\mathcal{M}$ $ \gets$ Users in group $j$ sorted in descending order of their utility values.}
    \State{NOU $(j) = \abs{\mathcal{M}}$}
    \State {$\beta \gets$ $\hat{Y}(j)$}
    \State {$\eta \gets$ 0}
    \While {$\beta \neq 0$ AND $\eta \leq $  NOU $(j) -1$  }
        \State $X(\mathcal{M}_{\eta}) = 1$
        \If {$X$ \text{is not feasible}} \label{line:feasCheck}
        \State $X(\mathcal{M}_{\eta}) = 0$
        \Else
        \State {$\beta = \beta -1$}
        \EndIf
        \State {$\eta = \eta+1 $}
        
\EndWhile 
\EndFor
\end{algorithmic}
\end{algorithm}

The number of groups $\abs{\hat{Y}}$ (i.e., the number of neurons in the output layer) is an important hyperparameter that can have a significant effect on the performance and accuracy of the model. Recall that $\abs{\hat{Y}}$ can be manipulated by varying $L$ and/or $V$. The accuracy of the model can be positively affected by the increase in the number of groups, since the search space of ~Alg. \ref{alg:cap} is reduced. Yet, an excessive number of groups (e.g., comparable to the number of EV users), hence output neurons, will likely result in highly sparse predictions leading to increased complexity and a higher chance of overfitting. These aspects combined can degrade the performance severely. Likewise, reducing the number of groups expands the search space thus requiring more computational power during post-processing. Consequently, adjusting this number can improve the  predictions made by the model, but it is important to consider the trade-offs between computational efficiency and performance. Ultimately, the optimal number of neurons in the output layer should be determined through empirical experimentation and evaluation using appropriate evaluation metrics.

To guarantee a feasible solution to {\sc EVCRP}, Alg.~\ref{alg:cap} verifies the constraints' violations in line \ref{line:feasCheck}. The computational complexity of this post-processing routine depends on the number of groups and the sorting procedure within each group. Assuming that the number of users in each group is constant (i.e., $<< |\cA|$), the computational time amounts to $\mathcal{O}\Big(|\hat{Y}| + |\cC|\cdot|\cT|\Big)$. In the unlikely worst-case scenario where {\sc DClEVerNet} forecasts a solution instance with all the users partitioned into a single group, the computational complexity would be $\mathcal{O}\Big(|\mathcal{A}| \log |\mathcal{A}| + |\cA|\cdot|\cC|\cdot|\cT|\Big)$.

\section{Evaluation studies}\label{sec:sims}
In this section, the performance of the proposed approach is evaluated by simulations. The considered criteria are the approximation ratio and the running time. 
\subsection{Simulation Setup and Settings}
\begin{figure}[!t]
     \centering
     \includegraphics[width=\linewidth]{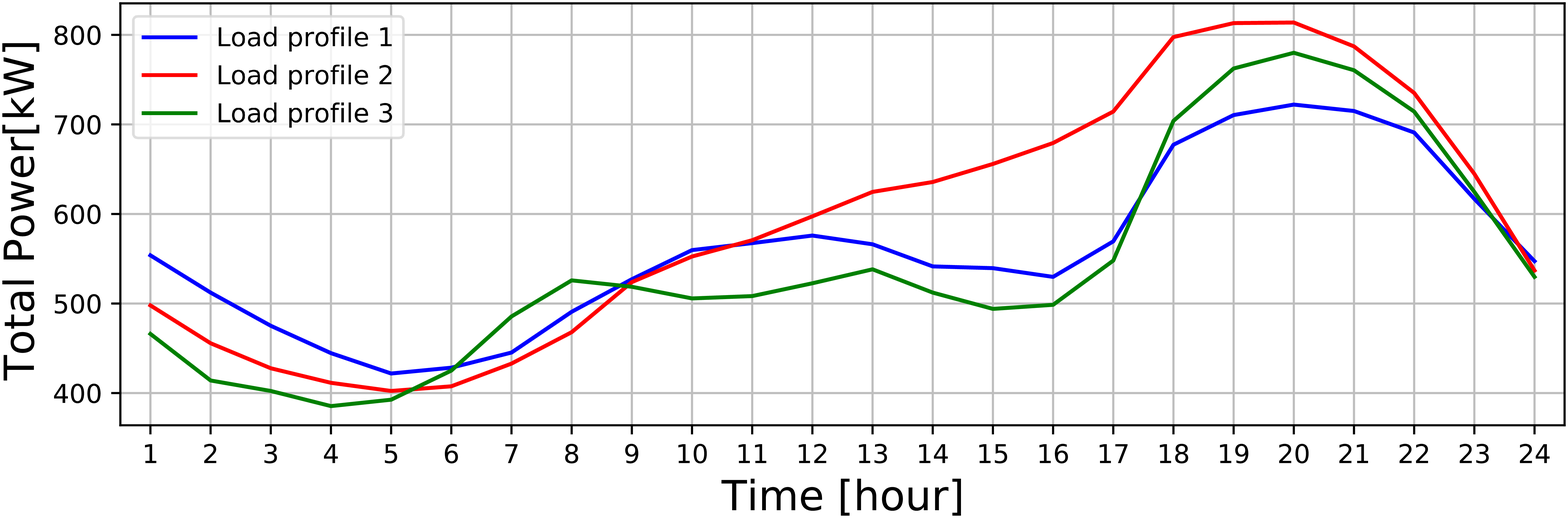}
    \caption{The considered three versions of base-load profiles based on $650$ households.}
     \label{fig:LoadProfiles}
 \end{figure}

\paragraph{Distribution network setting:}
The capacity of the power substation, denoted by $\overline{P}$, is set to be $1$MW. The base-load, denoted by $d(t)$, varies and is dependent on the time of the day. In our simulations, we adopt three base-load profiles as depicted in ~Fig. \ref{fig:LoadProfiles}; each profile  represents the daily average household load profile in the service area of South California from 00:00, January 1, 2011, to 23:59, January 3, 2011  \cite{sce_eca}. Each EV has the option to charge in $3$ charging stations with charging rates of $1.5k$W, $7$Kw, and $50$kW, respectively. In each station, the number of installed chargers, denoted by $N_c$, is set to $200$. 

\paragraph{Scheduling horizon:}
We consider a 24-hour time horizon divided into 96 slots of 15 minutes, i.e., $|\cT| =96$.
\paragraph{Training and Testing data:}
For training the presented {\sc DClEVerNet} model, $50,000 $input samples were produced, to which an $80$/$20$ split was applied to divide the data into training and test sets. The data is equally comprised of synthetic and real-world charging requests. The synthetic data are based on real-world patterns. According to \cite{qian2010modeling}, most EV users start charging when return home at 18:00, and more than $90$\% of EVs start charging between 13:00 and 23:00. Therefore, the start time can be modeled as a normal distribution with a mean of 18:00 and a standard deviation of $5$ hours. We set the charging start time at random (following a normal distribution) and its length to the time needed to fulfill the energy requirement. The energy required for  user $a \in \mathcal{A}$ was formulated as follows: 
 \begin{equation}
P^a=\Bigg(1-\frac{\text{SOC}^a}{100}\Bigg) \cdot B^a \,,
\end{equation}
where SOC$^a$ and $B^a $  are the state of charge and the battery size for user $a \in \mathcal{A}$.
 
The SOC is  modeled as a truncated normal distribution with a mean of $0.5$ and a standard deviation of $0.3$. For each EV, the SOC is limited between $20$\% to $80$\%. Similarly, $B^a$ is modeled as a normal distribution with a mean of $24$kW and a standard deviation of $10$kW, for $\forall a \in \cA$. 
For real-world data, we are utilizing the Caltech Adaptive Charging Network dataset~\cite{acn_dataset}, which is a comprehensive collection of data related to the energy consumption patterns and charging behaviors of electric vehicles within a network environment. The dataset contains rich information about EV driving and charging patterns. We adopt the arrival time and the power demand of EVs on Tuesdays. 

For inclusivity, this study evaluates two distinct settings for utility values. The first setting involves a linear utility, which can be expressed as:
\begin{equation}\label{eqn:LIN}
    u^a=0.36\cdot P^a \,,
\end{equation}
where 0.36 is the peak electricity price. The second setting involves a random selection of utilities within the range from $5000$ to $8000$.

\begin{figure*}
    \centering
    \begin{subfigure}{\columnwidth}
\includegraphics[width=1\columnwidth]{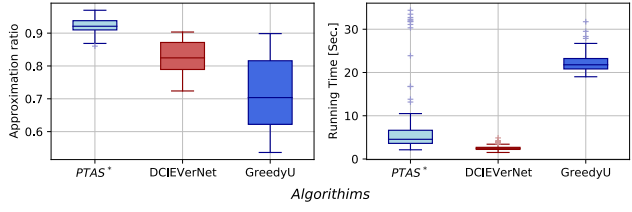}%
    \caption{The average approximation ratio and running time of PTAS$^*$, {\sc DclEVerNet} [Q=4, L=10,V=10], and GreedyU on $250$ {\sc EVCRP} instances randomly taken from the synthetic testing dataset.}%
    \label{subfiga}%
    \end{subfigure}\hfill%
    \begin{subfigure}{\columnwidth}
        \includegraphics[width=1\columnwidth]{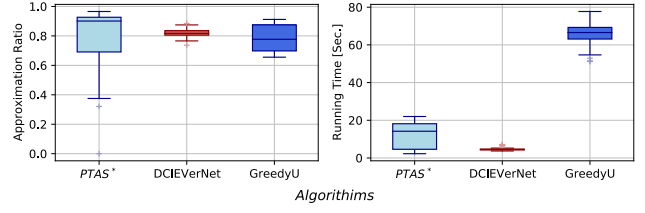}%
        \caption{The average approximation ratio and running time of PTAS$^*$, {\sc DclEVerNet} [Q=4, L=10,V=10], and GreedyU on $250$ {\sc EVCRP} instances randomly taken from the ACN testing dataset.}%
        \label{subfigb}%
    \end{subfigure}\hfill%
    \caption{Performance comparison of {\sc DclEVerNet}, GreedyU, and PTAS$^*$ algorithms on the synthetic and ACN datasets.}
    \label{fig:ALL-ALG}
\end{figure*}

In the case studies performed, the energy cost is based on  the time of use rate in South California ~\cite{sce_eca}. For both the training and testing data, the non-EV load is chosen to be the load profile~1 pictured in~Fig \ref{fig:LoadProfiles}.

\paragraph{Benchmark algorithms:}
\begin{itemize}
\item PTAS$^*$: A PTAS is an algorithm that for any $\epsilon > 0$ is guaranteed to produce a $(1-\epsilon)$-approximation to a given maximization problem. The running time of a PTAS is polynomial in the input size for every fixed $\epsilon$, but the exponent of the polynomial might depend on $\frac{1}{\epsilon}$. In other words, a PTAS allows trading approximation ratio for the running time. As one benchmark, we implement an adapted version of the PTAS for {\sc EVCRP} proposed in ~\cite{khonji2018approximation}. The adapted variant, denoted by PTAS$^*$, proceeds as follows. The algorithm considers randomly generated initial guesses for a small subset of users. For each guess, the remaining subproblem of {\sc EVCRP} is solved with a numerical solver by relaxing the other discrete control variables to be continuous, then rounding these variables to obtain a feasible solution. The key difference between PTAS and PTAS$^*$ is that the former iterates over all possible guess combinations in the search for the best approximation ratio, whereas the latter considers only a limited number of guesses. Even with a few users, the running time of PTAS in practice could be computationally prohibitive. Therefore, in PTAS$^*$, we restrict the number of guesses to $250$.   

\item Greedy Utility Algorithm (GreedyU): As a baseline, we adopt a commonly utilized greedy strategy that sorts the EV users in $\mathcal{A}=\{1, \hdots, \left|{\mathcal{A}}\right|\}$ according to their conditional gain in descending order, such that
\begin{equation}
    G^1 \geq G^2 \geq \hdots \geq G^{\left|{\mathcal{A}}\right|}\,,
\end{equation}
then selects the customers sequentially in that order, subject to feasibility constraints. Since GreedyU sorts and iterates over the entire customer set $\cA$, the running time complexity is $\mathcal{O}\Big(|\mathcal{A}| \log |\mathcal{A}| + |\cA|\cdot|\cC|\cdot|\cT|\Big)$.
\end{itemize}

\paragraph{Implementation of the DNN model:}
The presented {\sc DClEVerNet} model
was constructed on the basis of the Keras platform. The model was trained over $200$ epochs with a batch size of $32$. The learning rate was set to $0.001$.

The simulations were evaluated on a desktop machine with Intel i7-8750 CPU 2.2GHz processor and 16 GB of RAM. The algorithms were coded 
 in Python 2.7 programming language with SciPy and NumPy libraries for scientific computation.

\subsection{ Evaluation Results}

To thoroughly evaluate the effectiveness of the assembled {\sc DClEVerNet} model, we performed a series of experiments utilizing out-of-sample data from both synthetic and real-world data sets. In addition, in our evaluations, we  examine the generalizability of the trained model by testing its performance on out-of-distribution data. This involved manipulating various factors, such as reducing the number of chargers $N_c$, increasing the number of EV charging requests, and varying the load profile. Through these experiments, we aim to assess the method's ability to adapt and make accurate predictions in new and unseen situations.

 \begin{figure}[!t]
     \centering
     \includegraphics[trim={0.3cm 0cm 0.1cm 0cm},clip,width=\linewidth]{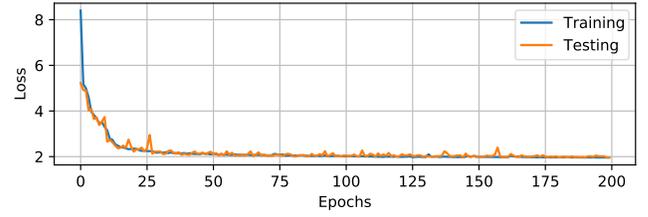}
    \caption{The evolution of training and validation losses over the number of epochs trained.}
     \label{fig:LOSSf}
 \end{figure}
 \begin{figure}[!b]
     \centering
     \includegraphics[trim={0.28cm 0cm 0.2cm 0cm},clip,width=\columnwidth]{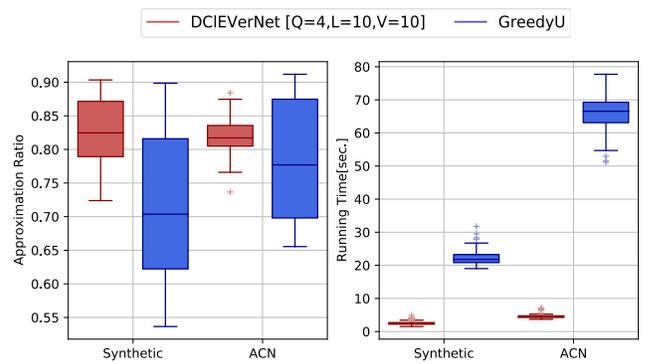}
    \caption{The average approximation ratio and running time of {\sc DclEVerNet} [Q=4, L=10,V=10] and GreedyU on $250$ {\sc EVCRP} instances randomly taken from the synthetic and ACN testing datasets.}
     \label{fig:DNN-GREEDY}
 \end{figure}

First, we analyze the behavior of the loss function over multiple epochs, as depicted in Fig.\ref{fig:LOSSf}. The results show a consistent decrease in the loss function for both the training and testing data. This indicates that the model is effectively learning to produce increasingly similar predictions to the corresponding labels. Next, we examine the approximation ratio (i.e., how close the predicted solution is to the optimal solution) and the running time of the model. The performance of {\sc DClEVerNet} was evaluated over $250$ testing instances from the synthetic and ACN data sets, respectively. The comparative results depicted in Fig.\ref{fig:DNN-GREEDY}, clearly demonstrate that {\sc DClEVerNet} outperforms the GreedyU algorithm in terms of both approximation ratio and computational time. In addition, a comparison was made between {\sc DClEVerNet} and PTAS$^*$, and the results, plotted in Fig.\ref{fig:ALL-ALG}, indicate that the average case performance of {\sc DClEVerNet} is nearly on par with that of PTAS$^*$ in terms of solution quality for both synthetic and ACN instances. As evidenced by Figs.\ref{fig:ALL-ALG} and \ref{fig:DNN-GREEDY}, for the ACN dataset the worst-case approximation ratio achieved by {\sc DClEVerNet}  was higher than those of PTAS$^*$ and the GreedyU method. Furthermore, {\sc DClEVerNet} demonstrated superior computational efficiency, on average being the fastest in all the case studies conducted.

\begin{figure}[!t]
     \centering
     \includegraphics[trim={0.28cm 0cm 0.2cm 0cm},clip,width=\columnwidth]{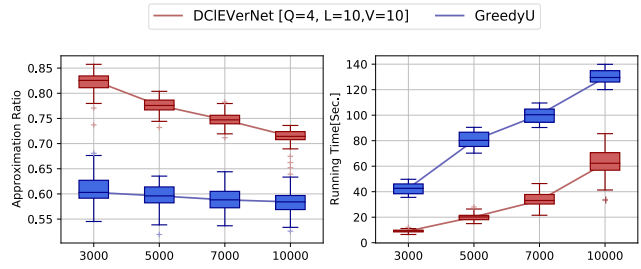}
    \caption{The average approximation ratio and running time of {\sc DclEVerNet} [Q=4, L=10,V=10], and GreedyU on $200$ randomly generated synthetic {\sc EVCRP} instances with gradually increasing input size from 3000 to 10,000.}
     \label{fig:Scalability}
 \end{figure}

To demonstrate the scalability of the proposed approach, we tested the trained {\sc DClEVerNet} model (on instances with $1500$ customers) on higher-dimensional problem instances with the number of EV requests ranging from $3,000$ to $10,000$. Each case was evaluated with $200$ synthetic instances with random utilities. The results are depicted in Fig.\ref{fig:Scalability}, which compares the performance of the proposed approach to that of the GreedyU algorithm. As observed from the figure, {\sc DClEVerNet} demonstrates a superior approximation ratio, and it's significantly faster than GreedyU across all instances. This indicates that the proposed approach can handle large-scale EV charging scheduling problems with acceptable performance without the need for re-training. 

\begin{figure}[!t]
    \centering
    \includegraphics[trim={0.28cm 0cm 0.2cm 0cm},clip,width=\columnwidth]{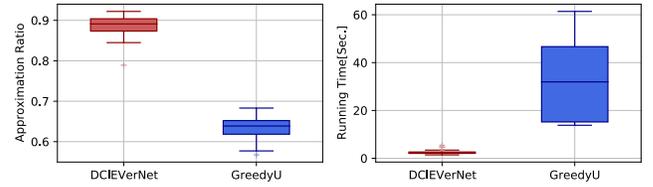}
    \caption{The average approximation ratio and running time of {\sc DclEVerNet} [Q=4, L=10,V=10] and GreedyU on $100$ {\sc EVCRP} instances considering modified number of charging slots ($N_c=100$).}
    \label{fig:less-slots}
\end{figure}
\begin{figure}[!t]
    \centering
    \includegraphics[trim={0.28cm 0cm 0.2cm 0cm},clip,width=\columnwidth]{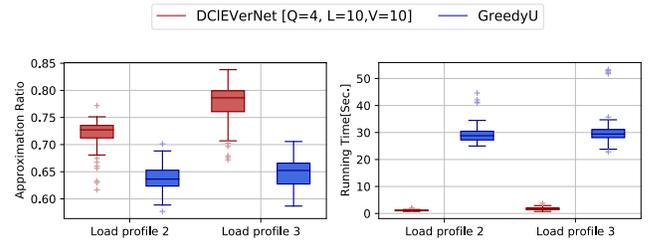}
    \caption{The average approximation ratio and running time of {\sc DclEVerNet} [Q=4, L=10,V=10] and GreedyU  on $200$ {\sc EVCRP} instances generated considering load profiles previously unseen during the training.}
    \label{fig:loadProfile}
\end{figure}

 The load profile is an important parameter in the EV charging scheduling problem as it affects the distribution of the energy demand, and thus it is crucial to evaluate the model's performance under different load profiles. The performance of {\sc DClEVerNet} was evaluated on $200$ instances with load profiles $2$ and $3$, which were not included in the training dataset. The results of this evaluation are presented in Fig.\ref{fig:loadProfile}. While the observed performance of the {\sc DClEVerNet} is lower than in the instances with load profile 1 (on which it was trained), it still significantly outperforms the GreedyU algorithm in both solution quality and running time. Lastly, to further assess the robustness and generalizability of the constructed model, we test the performance of the trained {\sc DClEVerNet} model on $100$ instances with a modified number of charging slots in each station: $N_c = 100$ for $\forall c \in \cC$. As Fig ~\ref{fig:less-slots} demonstrates, {\sc DClEVerNet} continued to perform superior to the GreedyU algorithm in terms of approximation ratio and computational efficiency. This signifies that the proposed approach is robust and can adapt to different load profiles while retaining satisfactory performance.

\section{Discussion and Prospective Extensions}\label{sec:discussion}

While the performed extensive simulations demonstrate the effectiveness of the trained sample \textsc{DclEVerNet} model, a number of further improvements can be attained to solidify the overall framework. We identify the following immediate extensions as well as prospective avenues for further exploration. 

\begin{enumerate}
    \item One promising immediate extension is to augment the proposed framework by incorporating more complex neural network architectures. Specifically, the utilization of Transformer based architectures, which employ the self-attention mechanism, has demonstrated remarkable performance in a wide range of applications. The self-attention mechanism enables the network to focus on specific parts of the input by assigning different weights to different parts of the input, this allows the network to focus on the most relevant parts of the input and improve the performance of the network. Another technique that can be considered is the integration of skip connections. These connections enable the flow of information to bypass certain layers in the network, which can improve the efficiency of the flow of information and improve the performance of the network. Combining these two and integrating them into \textsc{DclEVerNet} will allow training a substantially deeper network than the currently presented one, likely yielding drastically improved generalisability and performance.

    \item Another immediate extension is to substitute the employed post-processing routine. Depending on the desired outcome, one can improve the computational efficiency or the performance quality. For example, towards the latter, one can adopt a variant of the Beam Search algorithm. Beam Search is a heuristic exploration algorithm that maintains a set of candidate solutions based on different algorithms and then returns the one with the highest objective value. For example, instead of relying only on one sorting procedure, the post-processing can simultaneously employ several sorting criteria based on power demand and/or gain-to-power ratio, then return the selection of users with the highest objective value. This would allow to explore the predicted solution space more extensively. On the other hand, Alg.~\ref{alg:cap} can be replaced by a faster yet possibly less efficient procedure. One example of such a method is the selection of users probabilistically via a random process. In this worst-case scenario, this would consume only linear running time.

    \item In future work, the extension of this framework to include covering and mixed covering and packing problems will be explored. Covering problems involve finding a subset of items that can completely cover a set of resources, while mixed covering and packing problems involve both covering resources and packing items. The proposed framework can be extended to include these types of problems by creating a similar dataset as described in Sec.~\ref{sec:preprecoess}.

    \item Added to the latter, another promising direction is to completely eliminate the need for post-processing, thus providing an end-to-end DL optimizer for combinatorial optimization problems. Though profoundly difficult, we believe that it is plausible to devise a standalone DL solver (possibly a two-stage approach  combining two different DL techniques) which on expectation may return near-optimal approximately feasible (with a bounded violation error) solutions.  

\end{enumerate}

\section{Conclusion}\label{sec:conclusion}

This paper studied the problem of maximizing the total welfare gain of EV users participating in scheduling reservation programs in \textit{large-scale, networked} EV charging facilities. Aiming to inform and advance the design of scalable and efficient reservation management strategies, we introduced and empirically verified a Deep Combinatorial Learning pipeline that can attain near-optimal scheduling decisions within sub/near-linear time. To provide further scrutiny, we tested the constructed model's generalisability and robustness against out-of-sample and out-of-distribution inputs based on previously unseen settings and parameters of the problem. The findings signify the potential of the proposed approach to pave the way towards more efficient means of tackling combinatorial optimization problems with tens of thousands of decision variables.

\begin{acks}
We are very grateful to the anonymous referees for their time and effort spent on reviewing this work. Their valuable feedback and constructive comments helped to improve the paper's presentation quality and clarity. Also, we would like to extend our sincere appreciation to Anahit Sargsyan (a Ph.D. student at the Technical University of Munich) for reading, commenting on, and helping to polish the preliminary variant of this paper.
\end{acks}

\bibliographystyle{ACM-Reference-Format}
\bibliography{IEEEabrv, mybib}

\end{document}